\documentclass[10pt,twocolumn,letterpaper]{article}

\usepackage{cvpr}
\usepackage{times}
\usepackage{epsfig}
\usepackage{graphicx}
\usepackage{amsmath}
\usepackage{amssymb}
\usepackage{algorithm}
\usepackage{multirow}
\usepackage{mathtools,xparse}
\usepackage{subcaption}
\usepackage{siunitx}
\usepackage{stmaryrd}
\usepackage[noend]{algpseudocode}
\usepackage{placeins}
\usepackage{mdframed}

\usepackage{amsthm}

\algnewcommand\algorithmicforeach{\textbf{for each}}
\algdef{S}[FOR]{ForEach}[1]{\algorithmicforeach\ #1\ \algorithmicdo}

\newcommand{\trans}{\text{T}}
\newcommand{\matr}[1]{\mathbf{#1}}     

\newcommand{\Proj}{\matr{P}}
\newcommand{\Hom}{\matr{H}}
\newcommand{\Aff}{\matr{A}}
\newcommand{\Fund}{\matr{F}}

\newcommand{\Point}{\matr{p}}
\newcommand*\rot{\rotatebox{90}}

\DeclarePairedDelimiter{\norm}{\lVert}{\rVert}
\NewDocumentCommand{\normL}{ s O{} m }{%
  \IfBooleanTF{#1}{\norm*{#3}}{\norm[#2]{#3}}_{2}%
}

\usepackage[pagebackref=true,breaklinks=true,letterpaper=true,colorlinks,bookmarks=false]{hyperref}

\cvprfinalcopy 

\def\cvprPaperID{363} 

\begin{document}

\title{Five-point Fundamental Matrix Estimation for Uncalibrated Cameras}

\author{Daniel Barath\\
Machine Perception Research Laboratory\\
MTA SZTAKI, Budapest, Hungary\\
{\tt\small barath.daniel@sztaki.mta.hu}
}

\maketitle

\definecolor{schematic_figure_border_color}{rgb}{.7,.7,.7}
\mdfdefinestyle{schematic_figure}{%
	linewidth = 0.8pt,%
	linecolor = schematic_figure_border_color,%
    shadowsize = 0pt,
}

\begin{abstract}
	We aim at estimating the fundamental matrix in two views from five correspondences of rotation invariant features obtained by e.g.\ the SIFT detector. The proposed minimal solver first estimates a homography from three correspondences assuming that they are co-planar and exploiting their rotational components. Then the fundamental matrix is obtained from the homography and two additional point pairs in general position. The proposed approach, combined with robust estimators like Graph-Cut RANSAC, is superior to other state-of-the-art algorithms both in terms of accuracy and number of iterations required. This is validated on synthesized data and $561$ real image pairs. Moreover, the tests show that requiring three points on a plane is not too restrictive in urban environment and locally optimized robust estimators lead to accurate estimates even if the points are not entirely co-planar. As a potential application, we show that using the proposed method makes two-view multi-motion estimation more accurate.  
\end{abstract}

\section{Introduction}

\begin{figure}[htbp]
	\centering
	\includegraphics[width = 0.95\columnwidth]{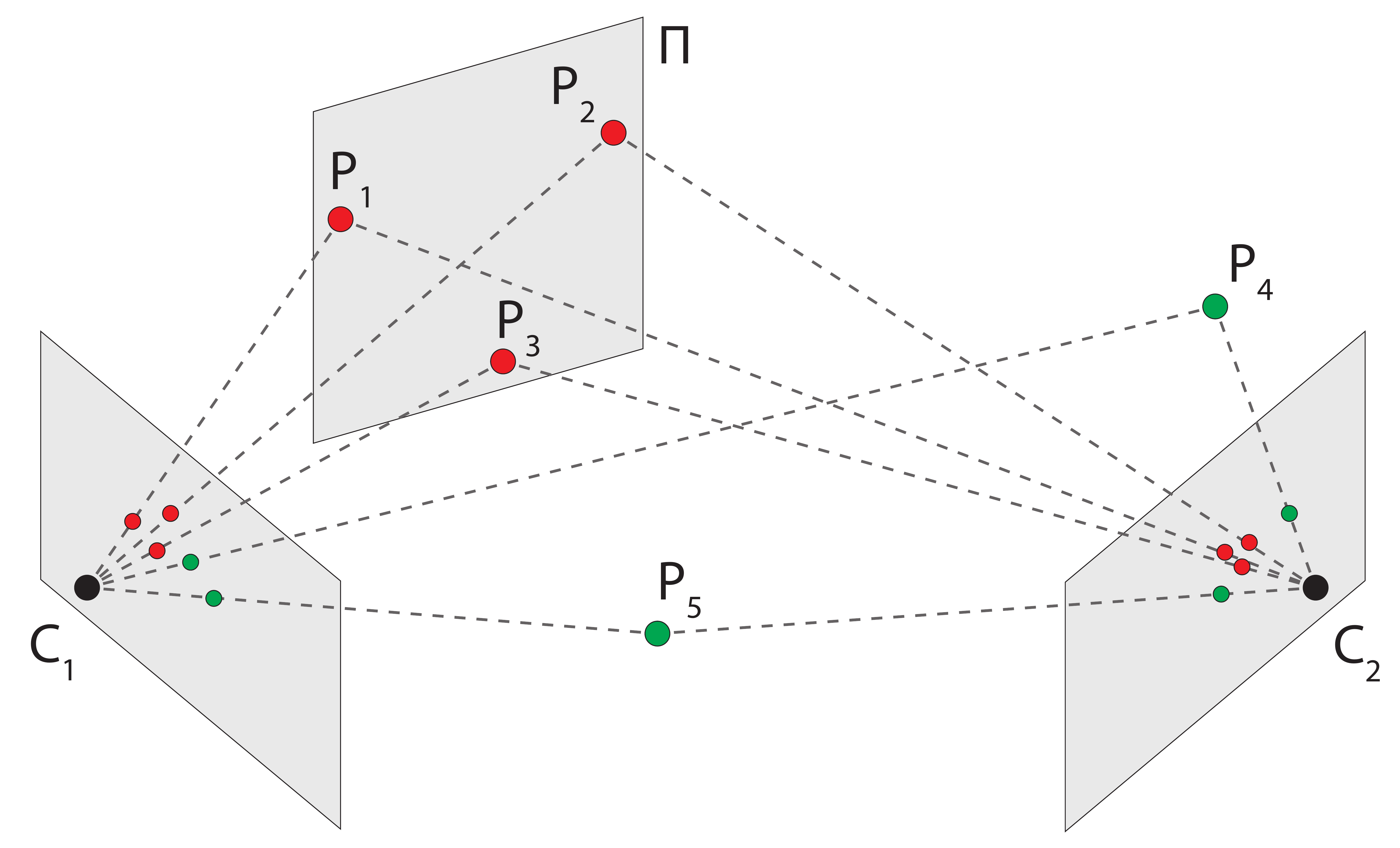}
	\caption{ The proposed minimal solver estimates a fundamental matrix between views $C_1$ and $C_2$. It first estimates a homography from three correspondences of co-planar points ($\textbf{P}_1$, $\textbf{P}_2$ and $\textbf{P}_3$) lying on plane $\pi$. The fundamental matrix is then obtained from the homography and two additional points ($\textbf{P}_4$ and $\textbf{P}_5$) in general position.}
	\label{fig:setup}
\end{figure}

This paper investigates the problem of estimating the relative motion of two \textit{non-calibrated cameras} from rotational invariant features. In particular, we are interested in the minimal case, i.e.\ to estimate fundamental matrix $\Fund \in \mathbb{R}^{3 \times 3}$ exploiting \textit{five} point correspondences together with rotational components obtained by, e.g.\ SIFT detector~\cite{lowe1999object}. The method requires three points to be co-planar and two additional ones in arbitrary position (see Fig.~\ref{fig:setup}). 

The classical way of estimating $\Fund$ for non-calibrated cameras is to apply the eight- or seven-point algorithms~\cite{hartley2003multiple}. They are both widely-used in the literature and fundamental tools of computer vision applications. The eight-point algorithm estimates the direct linear transformation induced by the epipolar constraint. The seven-point algorithm enforces the rank-two constraint by solving the cubic polynomial equation which it implies. 
From theoretical point of view, getting \textit{more information exclusively from point correspondences is not possible}. However, of course, there are approaches to reduce the number of unknowns. For example, knowing the intrinsic parameters of the cameras (i.e.\ the principal point, focal length, pixel ratios) enables to enforce the trace constraint. The problem becomes solvable using six point pairs~\cite{li2006simple,Kukelova2008BMVC,Stewenius2008IVC,torii2010six} if all intrinsics parameters but a common focal length are known, or five correspondences~\cite{nister2004efficient,li2006five,batra2007alternative,Kukelova2008BMVC,hartley2012efficient} are enough for fully calibrated cameras. One can also restrict the camera movement, e.g.\ the one point method proposed by Davide Scaramuzza~\cite{scaramuzza20111} assumes the cameras to move on a plane and the so-called non-holonomic constraint to hold.

By looking the other way, it is very rare nowadays to get solely the point coordinates from the applied feature detector. As an example, the widely-used SIFT detector provides a rotation and scale besides the coordinates. This additional information is rarely exploited in state-of-the-art geometric model estimators and just thrown away at the very beginning. \textit{This information is available} in most of the cases. In this paper, we aim at involving these additional \textit{affine parameters}, e.g.\ rotation of the feature, into the process to reduce the size of the minimal sample required for fundamental matrix estimation. 

Exploiting full affine correspondences (point correspondence, rotation, scales along both image axes and shear) for fundamental or essential matrix estimation, of course, is not a new idea. Perdoch et al.~\cite{PerdochMC06} proposed techniques for approximating the relative camera motion using two and three correspondences. Bentolila and Francos~\cite{Bentolila2014} proposed a method to estimate the exact, i.e.\ with no approximation, $\Fund$ from three correspondences. Raposo et al.~\cite{Raposo2016} proposed a solution for direct essential matrix estimation using two correspondences. 

Exploiting only a part of an affine correspondence, e.g.\ exclusively the rotation component, is a well-known technique for example in wide-baseline feature matching~\cite{matas2004robust}. However, to the best of our knowledge, the only work involving them into geometric model estimation is that of Barath et al.~\cite{barath2017phaf}. In \cite{barath2017phaf}, $\Fund$ is assumed to be known a priori and a technique is proposed for estimating a homography using two SIFT correspondences exploiting their scale and rotation components. Even so, an assumption is made, considering that the scales along axes $u$ and $v$ equal to that of the SIFT features -- which is generally not true in practice. Thus, the method yields only an approximation.

The contributions of the paper are: (i) we propose a technique for  estimating homography $\Hom$ using three rotation invariant feature correspondences. To recover $\Hom$, in addition to the point coordinates, the rotations of the features are exploited. (ii) The recovered homography is then used to calculate fundamental matrix $\Fund$ using two additional correspondences. (iii) It is reported on both synthesized and real worlds tests, that combining the proposed method with a robust estimator, e.g.\ LO-RANSAC~\cite{chum2003locally}, leads to results superior to the state-of-the-art in term of accuracy and the number of iterations required. Moreover, we demonstrate that using the proposed method in two-view multi-motion fitting is beneficial and leads to more accurate clusterings.

\section{Theoretical Background}

\paragraph{Affine Correspondences.}

In this paper, we consider an affine correspondence (AC) as a triplet: $(\Point_1, \Point_2, \Aff)$, where $\Point_1 = [u_1 \quad v_1 \quad 1]^\trans$ and $\Point_2 = [u_2 \quad v_2 \quad 1]^\trans$ are a corresponding homogeneous point pair in the two images (the projections of the 3D points in Fig.~\ref{fig:setup}), and 
\begin{equation*}
	\Aff = \begin{bmatrix}
		a_{1} & a_{2} \\
		a_{3} & a_{4}
	\end{bmatrix}
\end{equation*}
is a $2 \times 2$ linear transformation which we call \textit{local affine transformation}. To define $\Aff$, we use the definition provided in~\cite{Molnar2014} as it is given as the first-order Taylor-approximation of the $\text{3D} \to \text{2D}$ projection functions. Note that, for perspective cameras, $\Aff$ is the first-order approximation of the related \textit{homography} matrix
\begin{equation*}
	\Hom = \begin{bmatrix}
		h_1 & h_2 & h_3 \\
		h_4 & h_5 & h_6 \\
		h_7 & h_8 & h_9
	\end{bmatrix}
\end{equation*}
as follows: 
\begin{eqnarray}
		\begin{array}{llllll}
          a_{1} & = & \frac{\partial u_2}{\partial u_1} = \frac{h_{1} - h_{7} u_2}{s}, &
          a_{2} & = & \frac{\partial u_2}{\partial v_1} = \frac{h_{2} - h_{8} u_2}{s}, \\ 
          a_{3} & = & \frac{\partial v_2}{\partial u_1} = \frac{h_{4} - h_{7} v_2}{s}, &
          a_{4} & = & \frac{\partial v_2}{\partial v_1} = \frac{h_{5} - h_{8} v_2}{s}, 
		\end{array}
        \label{eq:taylor_approximation}
\end{eqnarray}
where $u_i$ and $v_i$ are the directions in the $i$th image ($i \in \{1,2\}$) and $s = u_1 h_7 + v_1 h_8 + h_9$ is the projective depth.

\paragraph{Fundamental matrix} 
\begin{equation*}
	\Fund = \begin{bmatrix}
		f_1 & f_2 & f_3 \\
		f_4 & f_5 & f_6 \\
		f_7 & f_8 & f_9
	\end{bmatrix}
\end{equation*}
is a $3 \times 3$ transformation matrix ensuring the so-called epipolar constraint $\Point_2^\trans \matr{F} \Point_1 = 0$ for rigid scenes. Since its scale is arbitrary and $\det(\Fund) = 0$, $\Fund$ has seven degrees-of-freedom (DoF). 
These properties will help us to recover the fundamental matrix from five rotation invariant feature correspondences.

\section{Homography from Three Correspondences}

In this section, it is shown how a homography can be estimated from three rotation invariant feature correspondences. First, we show the relationship of homographies and affine correspondences. Then this is decomposed into affine components establishing the way to exploit them independently. Selecting the appropriate equations from the obtained system, we finally use the given rotations to get the homography parameters. 

\subsection{Homographies and Affine Correspondences}

To form a linear equation system using $\Aff$, Eqs.~\ref{eq:taylor_approximation} are multiplied by the common denominator ($s$ -- projective depth), then rearranged as follows: 
\begin{eqnarray}
    \label{eq:orig_hom_aff}
	\begin{array}{ccc}
      h_1 - (u_2 + a_1 u_1) h_7 - a_1 v_1 h_8 - a_1 & = & 0 \\
      h_2 - (u_2 + a_2 v_1) h_8 - a_2 u_1 h_8 - a_2 & = & 0 \\
      h_4 - (v_2 + a_3 u_1) h_7 - a_3 v_1 h_8 - a_3 & = & 0 \\
      h_5 - (v_2 + a_4 v_1) h_8 - a_4 u_1 h_8 - a_4 & = & 0
	\end{array}
\end{eqnarray}
These equations encode the connection of a local affine transformation and a homography. 

As it is well-known, the relationship of a homography and a point correspondence $\Hom \Point_1 \sim \Point_2$ can be interpreted as an inhomogeneous linear system of equations. Note that operator $\sim$ means ``equality up to an arbitrary scale''. The system is as follows: 
\begin{eqnarray}
	\label{eq:dlt}
	\begin{array}{ccc}
    	u_1 h_1 + v_1 h_2 + h_3 - u_1 u_2 h_7 - v_1 u_2 h_8 & = & u_2 \\
    	u_1 h_4 + v_1 h_5 + h_6 - u_1 v_2 h_7 - v_1 v_2 h_8 & = & v_2
	\end{array}
\end{eqnarray}

Combining Eqs.~\ref{eq:orig_hom_aff} and \ref{eq:dlt}, an affine correspondence yields six linear equations on total. Thus each of them reduces the DoF of homography estimation by six. 

\paragraph{Affine Transformation Model.}

Although the relationship of full affine correspondences and homographies are well-defined, the current problem is the exploitation of features containing only a part of $\Aff$ -- the rotation. Therefore, let us define an affine transformation model as a combination of linear transformations as follows: 
\begin{eqnarray}
	\label{eq:affine_model}
    \begin{array}{lcl}
    	\Aff & = & \begin{bmatrix}
          a_{1}  & a_{2}  \\
          a_{3}  & a_{4} 
   		\end{bmatrix} = 
   		\begin{bmatrix}
          \cos(\alpha) & -\sin(\alpha) \\
          \sin(\alpha) & \cos(\alpha)
   		\end{bmatrix} 
   		\begin{bmatrix}
          s_{u} & w \\
          0 & s_{v}
   		\end{bmatrix} = \\[5mm]
   		& & \begin{bmatrix}
          s_{u} \cos(\alpha) & w \cos(\alpha) - s_{v} \sin(\alpha) \\
          s_{u} \sin(\alpha) & w \sin(\alpha) + s_{v} \cos(\alpha)
   		\end{bmatrix},
    \end{array}
\end{eqnarray}
where $\alpha$, $s_{u}$, $s_{v}$, and $w$ are the rotational angle, scales along axes $u$ and $v$, and shear parameter, respectively. 


Substituting the components of the matrix defined in Eqs.~\ref{eq:affine_model} into Eqs.~\ref{eq:orig_hom_aff}, the following system is given: 
\begin{eqnarray}
	\label{eq:hom_aff_sys}
    \begin{array}{lcl}
    	h_1 - u_2 h_7 - u_1 c_{\alpha} s_{u} h_7 - v_1 c_{\alpha} s_{u} h_8 - c_{\alpha} s_{u} & = & 0, \\
        h_2 - u_2 h_8 + v_1 c_{\alpha} w h_8 - v_1 s_{\alpha} s_{v} h_8 - & & \\
        u_1 c_{\alpha} w h_8 + u_1 s_{\alpha} s_{v} h_8 - c_{\alpha} w + s_{\alpha} s_{v} & = & 0, \\
        h_4 - v_2 h_7 - u_1 s_{\alpha} s_{u} h_7 - v_1 s_{\alpha} s_{u} h_8 - s_{\alpha} s_{u} & = & 0, \\
        h_5 - v_2 h_8 - v_1 s_{\alpha} w h_8 - v_1 c_{\alpha} s_{v} h_8 - & & \\
        u_1 s_{\alpha} w h_8 - u_1 c_{\alpha} s_{v} h_8 - s_{\alpha} w - c_{\alpha} s_{v} & = & 0,
    \end{array}   
\end{eqnarray}
where $c_{\alpha} = \cos(\alpha)$ and $s_{\alpha} = \sin(\alpha)$. Note that this system shows the general way of the affine parameters affecting the related homography. Even though we will consider exclusively $\alpha$ to be known in the subsequent sections, one can easily exploit these equations to solve for different features containing e.g.\ scales or shear besides the rotation. 

\subsection{Homography Estimation}

Assume three co-planar point correspondences $\Point_{1,i} = [u_{1,i} \quad v_{1,i} \quad 1]^\trans$, $\Point_{2,i} = [u_{2,i} \quad v_{2,i} \quad 1]^\trans$ ($i \in [1,3]$) and the related rotation components $\alpha_i$, obtained by e.g.\ SIFT, to be known. The objective is to find homography $\Hom$ for which $\Hom \Point_{1,i} \sim \Point_{2,i}$ and also satisfies Eqs.~\ref{eq:hom_aff_sys}. 

In the first part of the algorithm, only the coordinates are used to reduce the number of unknown parameters. We form $\Hom \Point_{1,i} \sim \Point_{2,i}$ (Eq.~\ref{eq:dlt}) for all correspondences as a homogeneous linear system $\textbf{B} \textbf{h} = 0$. Since each point pair yields two equations for the nine unknowns, coefficient matrix $\textbf{B}$ is of size $6 \times 9$ and $\textbf{h} = [h_1 \; h_2 \; h_3 \; h_4 \; h_5 \; h_6 \; h_7 \; h_8 \; h_9]^\trans$ is the vector of unknown parameters. The null-space of $\textbf{B}$ is three-dimensional, therefore the final solution is calculated as a linear combination of the three null-vectors as follows:
\begin{equation}
	\label{eq:null_space}
	\textbf{h} = \beta \textbf{b} + \gamma \textbf{c} + \delta \textbf{d},
\end{equation}
where $\textbf{b} = [b_1 \; ... \; b_9]^\trans$, $\textbf{c} = [c_1 \; ... \; c_9]^\trans$ and $\textbf{d} = [d_1 \; ... \; d_9]^\trans$ are the null-vectors, and $\beta$, $\gamma$, $\delta$ are unknown scalars. Due to the scale ambiguity of $\Hom$ one of them can be set to an arbitrary value, thus in our algorithm, $\delta = 1$.

Remember, that three rotation components are given, each providing four equations and three unknowns via Eqs.~\ref{eq:hom_aff_sys}. Two rotations yield eight equations and six unknowns, therefore, they are enough for estimating $\beta$ and $\gamma$. To exploit them, Eqs.~\ref{eq:null_space} have to be substituted into Eqs.~\ref{eq:hom_aff_sys} replacing each $h_j$ by $\beta b_j + \gamma c_j + d_j$ ($j \in [1,9]$). Since the scale along axis $v$ and shear $w$ are not known, the 2nd and 4th equations of Eqs.~\ref{eq:hom_aff_sys} yield no additional information, they are removed from the system. Without them, the two rotations lead to a multivariate polynomial system consisting of four equations with monomials 
$[\beta \quad \gamma \quad s_{u,1} \quad s_{u,1} \beta \quad s_{u,1} \gamma \quad s_{u,2} \quad s_{u,2} \beta \quad s_{u,2} \gamma]^\trans$.
Coefficient matrix $\textbf{C}$ is visualized in Table~\ref{tab:coefficient_mat}.
\begin{table}
    \caption{Homography estimation. Coefficient matrix $\textbf{C}$ of the multivariate polynomial system to which the rotation components lead. Each column represents the coefficients of a monomial (1st row) in the four equations (rows).}
  	\resizebox{.99\columnwidth}{!}{\begin{tabular}{c c c c c c c c }
    	$\beta$ & $\gamma$ & $s_{u,1}$ & $s_{u,1} \beta$ & $s_{u,1} \gamma$ & $s_{u,2}$ & $s_{u,2} \beta$ & $s_{u,2} \gamma$ \\ 
      	\hline
        $c_{11}$ & $c_{12}$ & $c_{13}$ & $c_{14}$ & $c_{15}$ & $c_{16}$ & $c_{17}$ & $c_{18}$ \\
    	& & & \multicolumn{2}{c}{...}  & & & \\
        $c_{41}$ & $c_{42}$ & $c_{43}$ & $c_{44}$ & $c_{45}$ & $c_{46}$ & $c_{47}$ & $c_{48}$ \\
      	\hline
  \end{tabular}}
	\label{tab:coefficient_mat}
\end{table}
Since four equations are given for four unknowns ($s_{u,1}$, $s_{u,2}$, $\beta$, and $\gamma$), and there are no higher order monomials, the system can straightforwardly be rearranged, then solved. The final formulas for $\beta$ and $\gamma$ are shown in Appendix~\ref{app:homography_parameters}. Finally, homography $\Hom$ is recovered through Eq.~\ref{eq:null_space}. 

Note that assuming that close points more likely belong to the same homography, we choose the rotations of the two closest points. Although this is a heuristics, it worked well in our experiments and does not require much computation. For problems, where the time is not critical, it is a possible choice to estimate the three homographies which the three rotations induce and select the one with the most inliers. 
Also note that all minimal samples, i.e.\ the selected five correspondences, can be rejected for which the two points in general positions also lie on the plane, thus leading to degenerate configuration. This can be checked by simply thresholding the re-projection error implied by $\Hom$ and each point pair.  

\section{Fundamental Matrix Estimation from \\Five Correspondences}

Suppose that homography $\Hom$, estimated in the previously described way, and two additional point correspondences are given. The objective is to estimate fundamental matrix $\Fund$
%
%
compatible both with $\Hom$ and the two correspondences and $\det(\Fund) = 0$ holds. The compatibility with $\Hom$ could be ensured through the well-known formula~\cite{hartley2003multiple}: 
%
$	\Hom^\trans \Fund + \Fund^\trans \Hom = 0 $.
%
However, the \textit{direct linear method} solving this system is unstable for inaccurate homographies, sometimes leading to completely meaningless results. The reason is that the samples are far from the normal distribution required for least squares fitting to work reasonably well~\cite{szeliski1998geometrically}. Zhou et al.~\cite{zhou2015revisit} proposed a normalization technique solving this problem, even so, this method needs at least three homographies to be known and do not consider the case when additional correspondences are given. Thus we chose the \textit{hallucinated point} technique generating five point correspondences using $\Hom$. The five generated and two given point pairs yield seven linear equations through $\Point_{2,i}^\trans \Fund \Point_{1,i} = 0$ ($i \in [1,7]$). Combining them, the following homogeneous linear system is given: 
$ 
	\textbf{D} \textbf{f} = 0,
$ 
where $\textbf{D}$ is the coefficient matrix and $\textbf{f} = [f_1 \; f_2 \; f_3 \; f_4 \; f_5 \; f_6 \; f_7 \; f_8 \; f_9]^\trans$ is the vector of unknown parameters. Matrix $\textbf{D}$ is as 
\begin{eqnarray*}
	\mathbf{D} = \\
    \resizebox{.99\hsize}{!}{$\begin{bmatrix} 
        u_{1,1} u_{2,1} & v_{1,1} u_{2,1} & u_{2,1} & u_{1,1} v_{2,1} & v_{1,1} v_{2,1} & v_{2,1} & u_{1,1} & v_{1,1} & 1 \\ 
		 &  &  & \multicolumn{2}{c}{...} &  &  &  &  \\
        u_{1,7} u_{2,7} & v_{1,7} u_{2,7} & u_{2,7} & u_{1,7} v_{2,7} & v_{1,7} v_{2,7} & v_{2,7} & u_{1,7} & v_{1,7} & 1 
	\end{bmatrix}$}.
\end{eqnarray*}
Note that making the estimator more stable, the normalization proposed by Hartley~\cite{hartley1997defense} is applied and the equations from the three co-planar points are also added. 
The null-space of matrix $\matr{D}$ is two-dimensional and the solution is calculated as the linear combination of the two null-vectors:
\begin{equation}
	\Fund =  \epsilon \matr{e} + \eta \matr{g}, 
    \label{eq:fund_null_space}
\end{equation}
where $\epsilon$ and $\eta$ are unknown scalars, $\matr{e} = [e_1\; ...\; e_9]^\trans$ and $\matr{g} = [g_1\; ...\; g_9]^\trans$ are the null-vectors. Due to the scale ambiguity of $\Fund$, $\eta$ can be set to an arbitrary value. To achieve stability we use $\eta = 1 - \epsilon$, thus keeping the sum of the weights to be one. Substituting Eq.~\ref{eq:fund_null_space} into $\det(\Fund) = 0$ leads to a cubic polynomial equation. The possible solutions for $\epsilon$ (their number is $\in \{1,2,3\}$, similarly to the seven-point algorithm) are obtained as the real roots of the polynomial. The resulting fundamental matrices are finally calculated by substituting each $\epsilon$ to Eq.~\ref{eq:fund_null_space}. 
Note that all fundamental matrices are discarded for which the \textit{oriented} epipolar constraint~\cite{chum2004epipolar} does not hold. 

Concluding the current and the previous sections, fundamental matrix $\Fund$ can be estimated from three co-planar and two arbitrary correspondences of rotation invariant features. 
 
\section{Experimental Results}

In this section, we compare the proposed method with the widely used seven- and eight-point algorithms~\cite{hartley2003multiple} both on synthesized and real worlds tests. 
\textit{The Matlab implementation is submitted as supplementary material.}

\subsection{Synthesized Tests}
\begin{figure*}[htbp]
	\centering
    \includegraphics[width = 0.65\columnwidth]{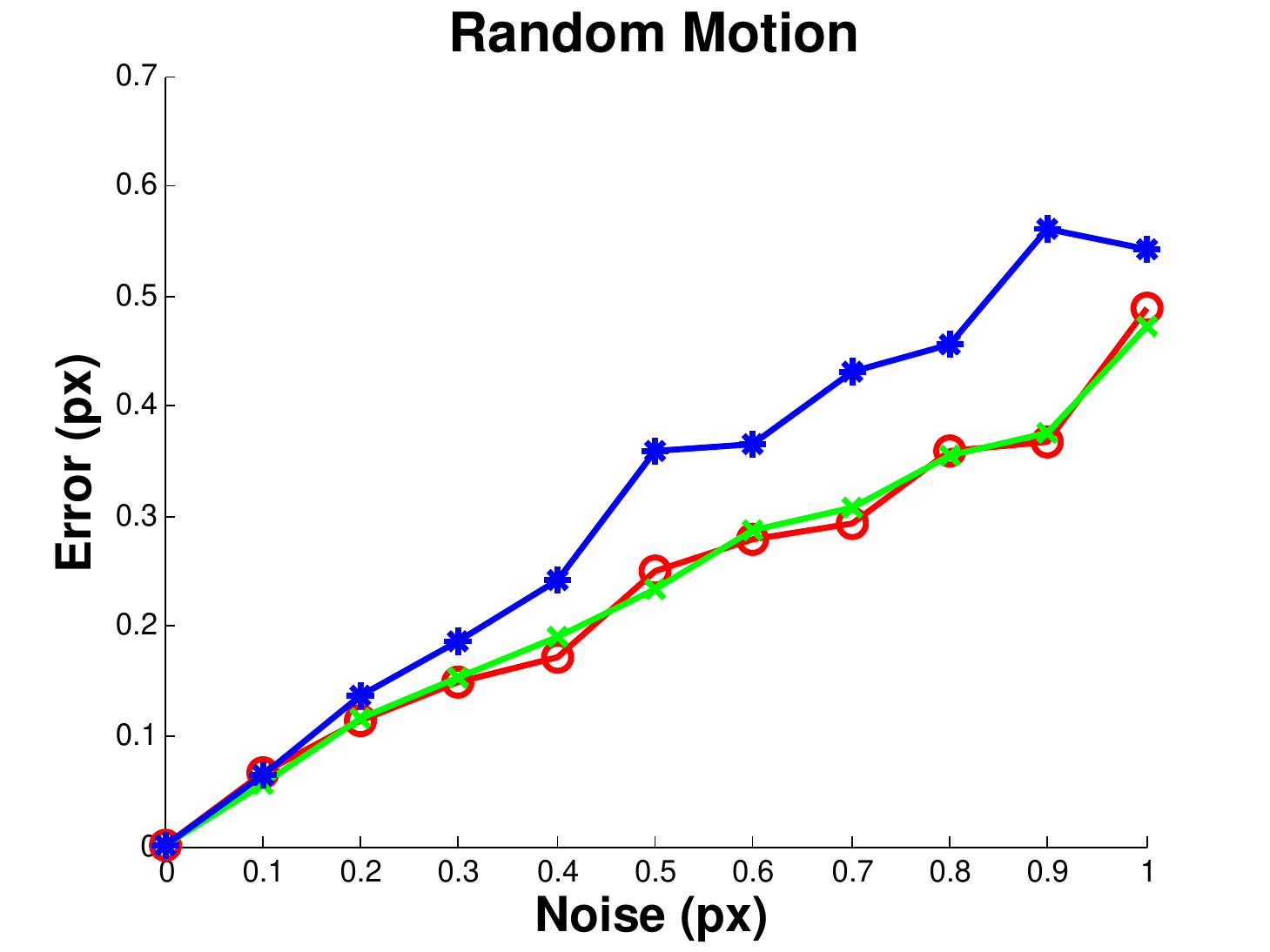}
    \includegraphics[width = 0.65\columnwidth]{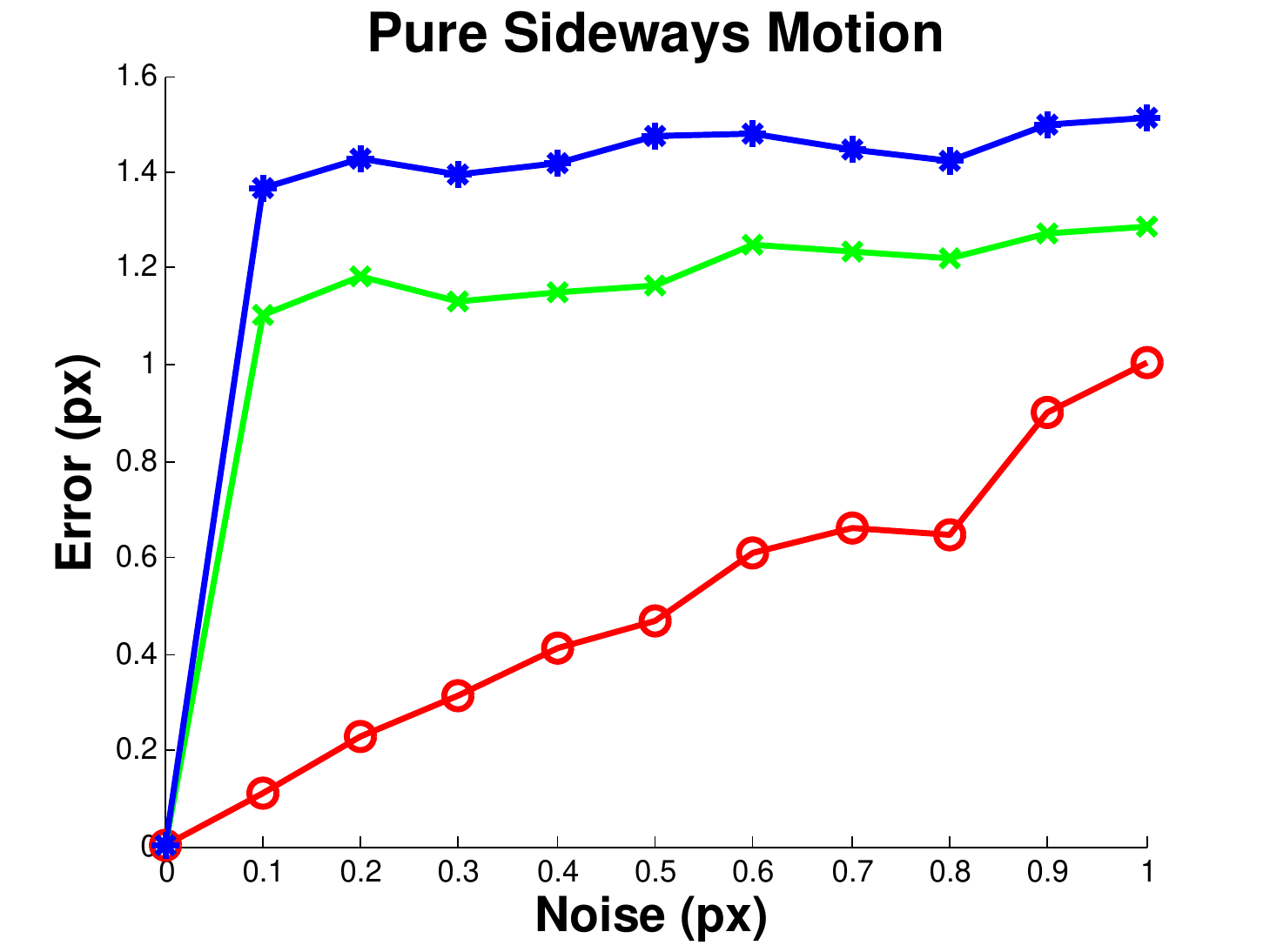}
    \includegraphics[width = 0.65\columnwidth]{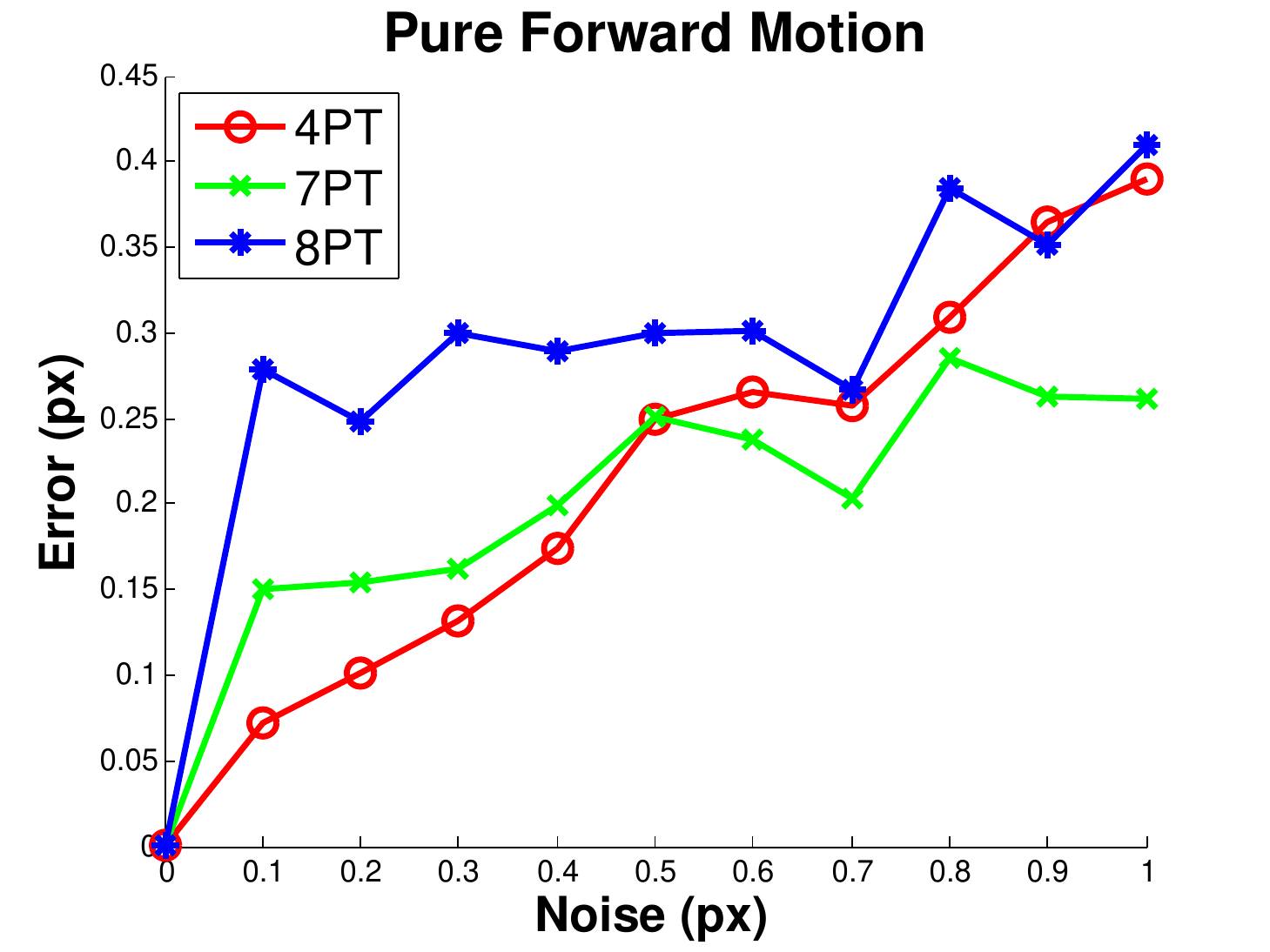}
	\caption{ The mean error (in pixels; plotted as the function of the noise $\sigma$) of the proposed, seven- and eight-point algorithms on cameras motions: random (left), sideways (middle) and forward (right). For random motion, both cameras are placed at a random point of a $10$-radius sphere and look towards the origin. For sideways and forward motions, the distance of the cameras was $10$ unit and a small zero-mean Gaussian-noise (with standard deviation set to $0.1$) is added to each coordinate. }
	\label{fig:synthetic_tests}
\end{figure*}

For synthesized testing, two perspective cameras were generated by their projection matrices $\Proj_1, \Proj_2 \in \mathbb{R}^{3 \times 4}$ and five random planes were sampled, each at four locations. The generated $20$ points were then projected onto the cameras and the ground truth affine transformations were computed from the image points and plane parameters. Zero-mean Gaussian-noise were added to the point coordinates, thus contaminating the affine components as well. 

Fig.~\ref{fig:synthetic_tests} shows the results of the proposed, eight- and seven-points algorithms applied to view pairs with specific camera motions (left -- random motion, middle -- pure sideways motion, right -- pure forward motion). The error is plotted as the function of the noise $\sigma$ (horizontal axis; in pixels). It is the mean symmetric epipolar distance from the correspondences not used for the estimation. For random motion, both cameras were located at a random point of a $10$-radius sphere and look towards the origin. For sideways and forward motions, the distance of the cameras was $10$ unit and a small perturbation, i.e.\ zero-mean Gaussian-noise with $0.1$ standard deviation, was added to the camera coordinates. 

It can be seen, that the proposed method leads similar accuracy to the seven-point algorithm for general movement. However, for purely sideways motion, the method is significantly less sensitive to the noise than the other competitors. For forward motion, if the noise $\sigma$ does not exceed $0.5$, the $5$-point technique is most accurate. After that point, the seven-point algorithm outperforms it.  

\subsection{Real World Tests}

\begin{figure}[htbp]
	\centering
    \includegraphics[width = 0.49\columnwidth]{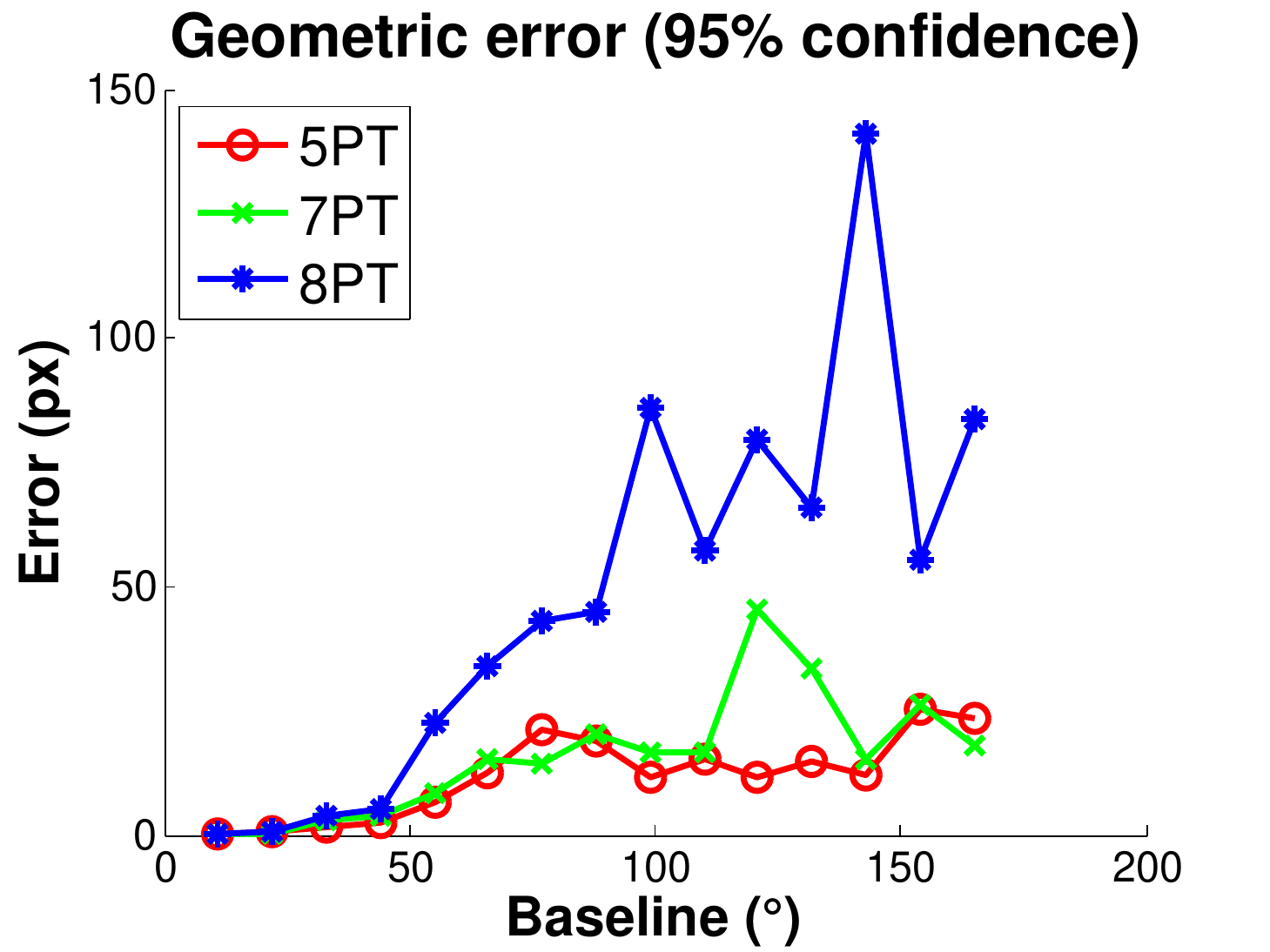}
    \includegraphics[width = 0.49\columnwidth]{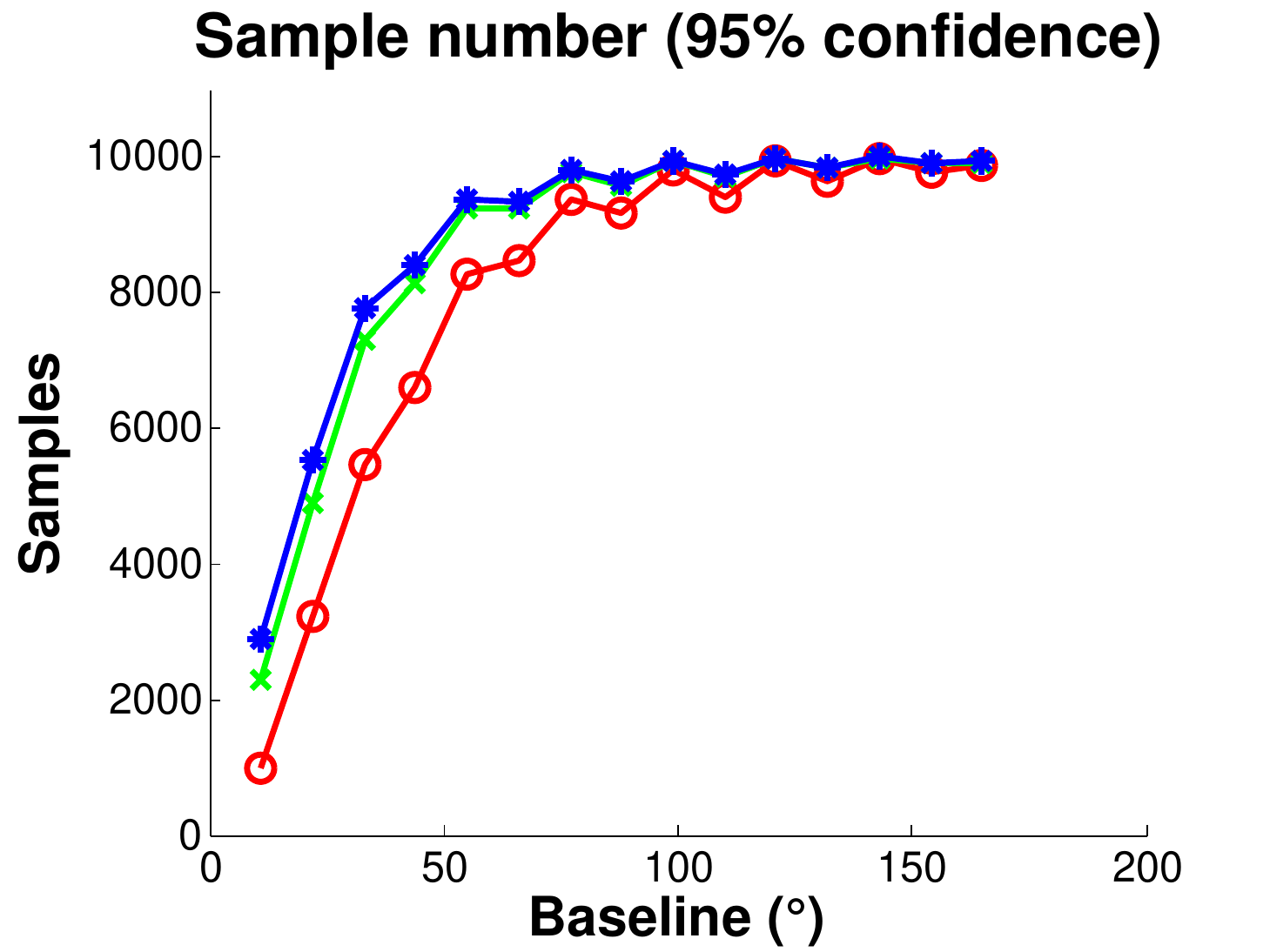}\\[0.2cm]
    \includegraphics[width = 0.49\columnwidth]{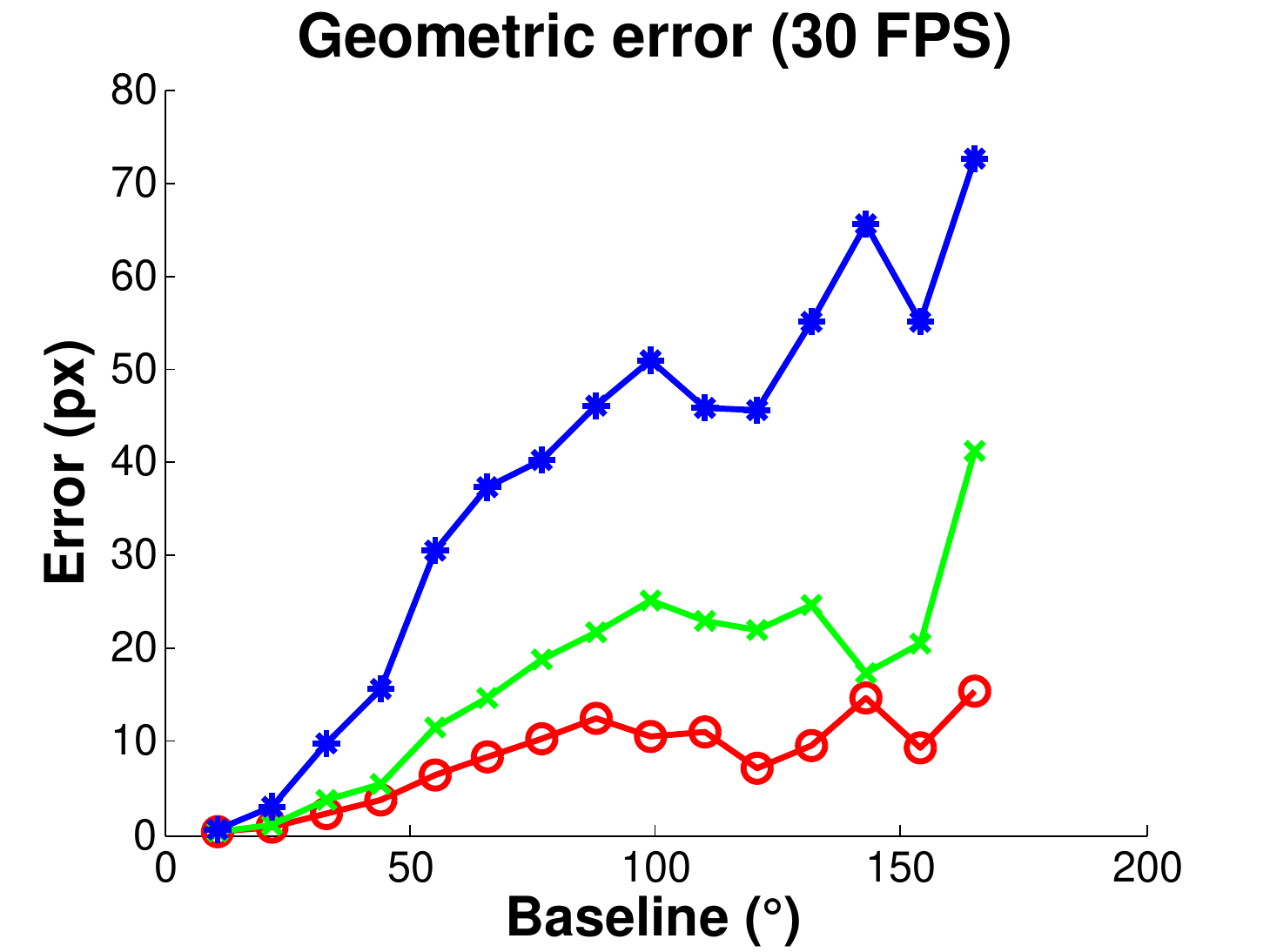}
    \includegraphics[width = 0.49\columnwidth]{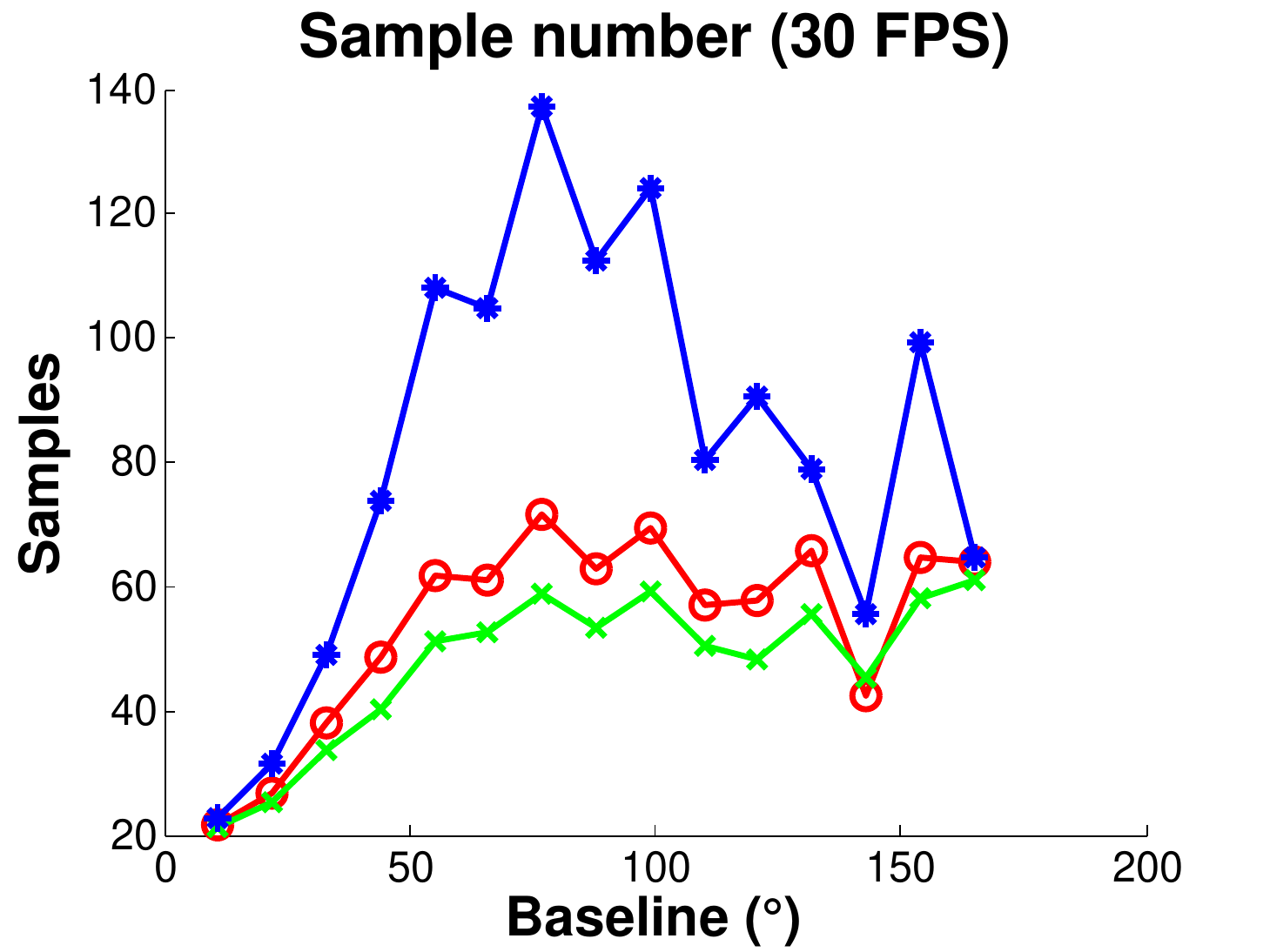}
	\caption{ The mean error (left; in pixels) and sample number (right) plotted as the function of the baseline (in degrees; rotation around the object) for confidence $99\%$ (top) and time limit $1 / 30$ secs (bottom). Results are computed from 100 runs on each image pair (\#515) in the {\fontfamily{cmtt}\selectfont Strecha} dataset. }
	\label{fig:increasing_baseline}
\end{figure}

\begin{figure}[htbp]
	\centering
    \begin{subfigure}{0.99\columnwidth}
		\includegraphics[width = 0.49\columnwidth]{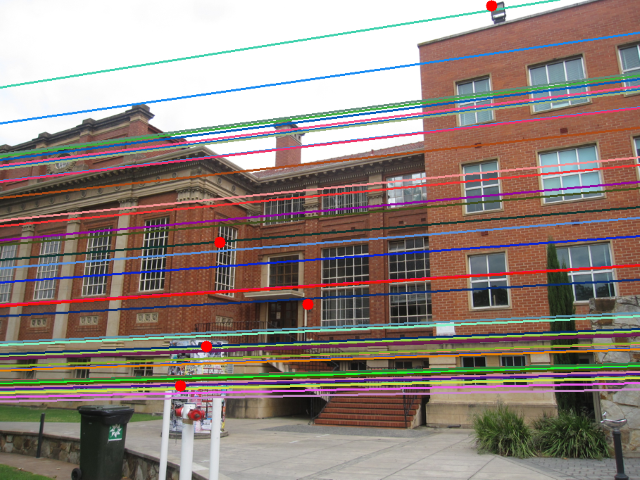}
		\includegraphics[width = 0.49\columnwidth]{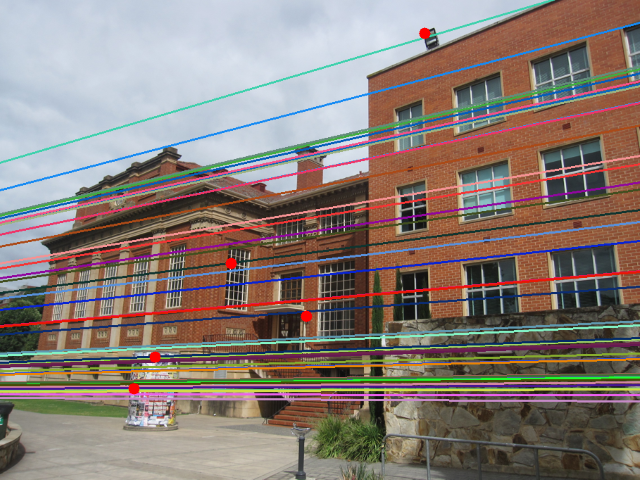}
		\caption{ {\fontfamily{cmtt}\selectfont AdelaideRMF} }
    \end{subfigure}
    \begin{subfigure}{0.99\columnwidth}
		\includegraphics[width = 0.49\columnwidth]{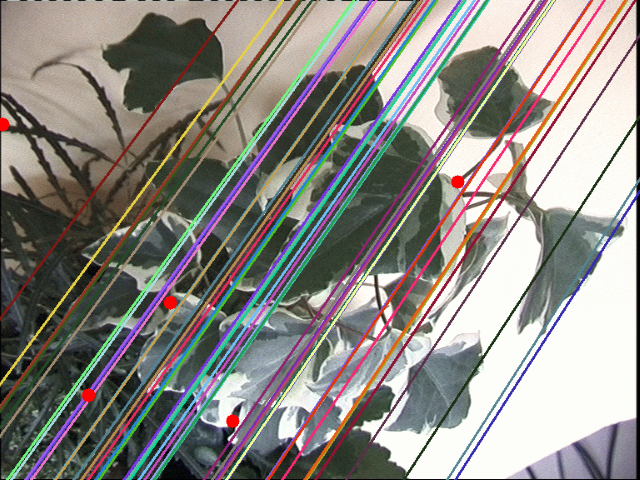}
		\includegraphics[width = 0.49\columnwidth]{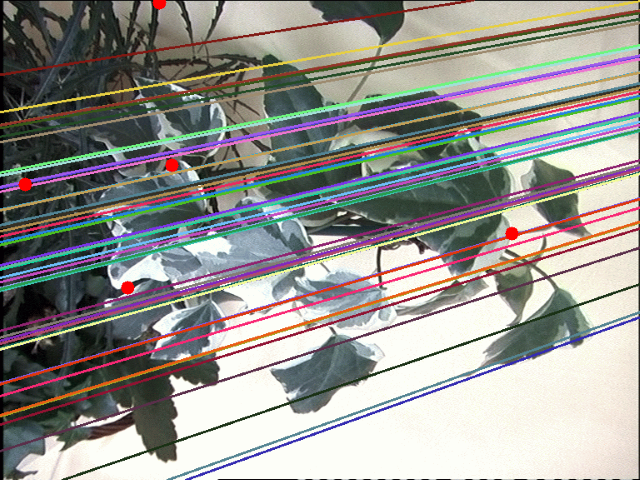}
		\caption{ {\fontfamily{cmtt}\selectfont Kusvod2} }
    \end{subfigure}
    \begin{subfigure}{0.99\columnwidth}
		\includegraphics[width = 0.49\columnwidth]{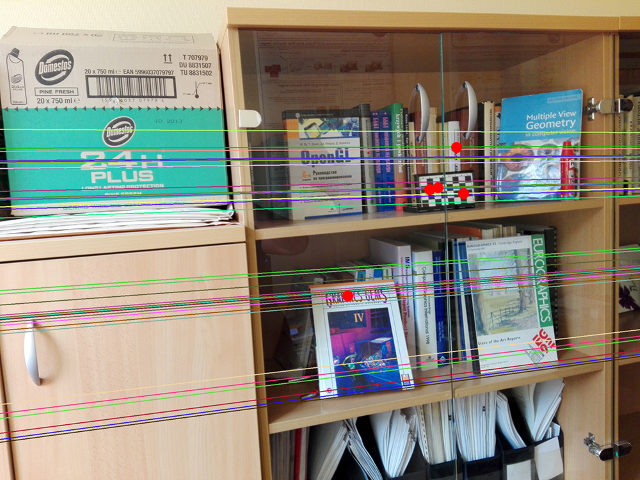}
		\includegraphics[width = 0.49\columnwidth]{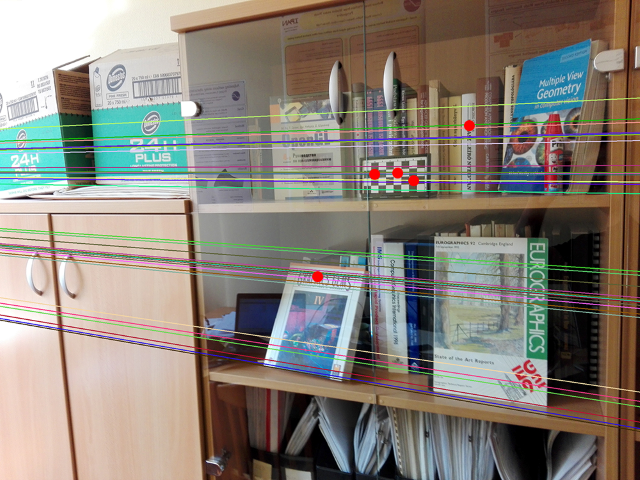}
		\caption{ {\fontfamily{cmtt}\selectfont Multi-H} }
    \end{subfigure}\\
    \begin{subfigure}{0.99\columnwidth}
		\includegraphics[width = 0.49\columnwidth]{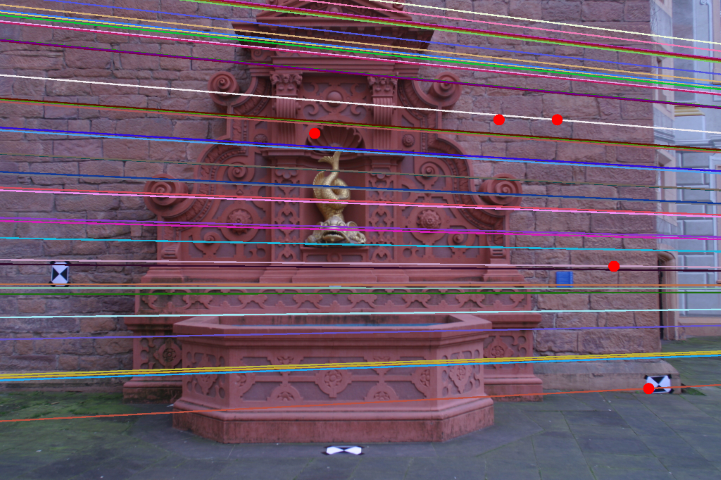}
		\includegraphics[width = 0.49\columnwidth]{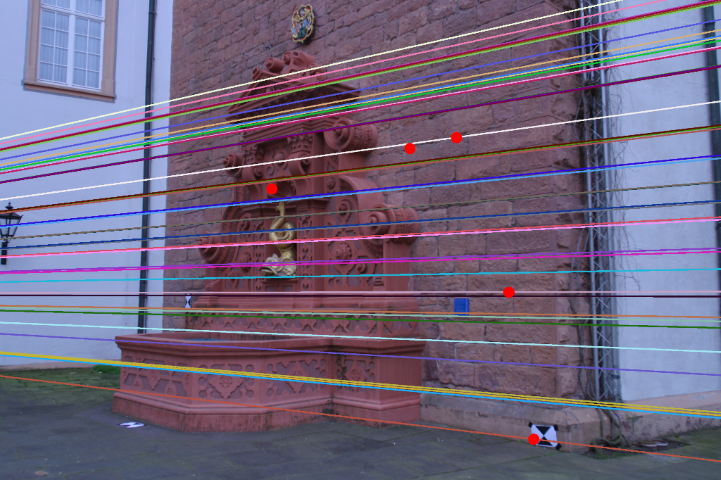}
		\caption{ {\fontfamily{cmtt}\selectfont Strecha} }
    \end{subfigure}
	\caption{ The results of the proposed method combined with Graph-Cut RANSAC. An image pair from each dataset with the corresponding epipolar lines of $50$ random inliers drawn by colors. The five point pairs which are used as the minimal sample are visualized by red dots. }
	\label{fig:real_images}
\end{figure}

To test the proposed method on real world data, we used the {\fontfamily{cmtt}\selectfont AdelaideRMF}\footnote{\url{cs.adelaide.edu.au/~hwong/doku.php?id=data}}, {\fontfamily{cmtt}\selectfont Kusvod2}\footnote{\url{cmp.felk.cvut.cz/data/geometry2view}}, {\fontfamily{cmtt}\selectfont Multi-H}\footnote{\url{web.eee.sztaki.hu/~dbarath}}, and {\fontfamily{cmtt}\selectfont Strecha}\footnote{\url{cvlab.epfl.ch/data/strechamvs}} datasets (see Fig.~\ref{fig:real_images} for examples). {\fontfamily{cmtt}\selectfont AdelaideRMF}, {\fontfamily{cmtt}\selectfont Kusvod2} and {\fontfamily{cmtt}\selectfont Multi-H} consist of image pairs of resolution from $455 \times 341$ to $2592 \times 1944$ and manually annotated (assigned to outlier or inlier classes) correspondences. Since the reference points do not contain rotation components we detected and matched points applying SIFT detector.

{\fontfamily{cmtt}\selectfont Strecha} dataset consists of image sequences (each image is of size $3072 \times 2048$) and a projection matrix for every image. Therefore, we paired the images in each sequence in every possible way. The ground truth $\Fund$ was estimated from the projection matrices~\cite{hartley2003multiple} and SIFT was used to get correspondences. Every detected point pair was considered as a reference point for which the symmetric epipolar distance~\cite{hartley2003multiple} from the ground truth $\Fund$ was smaller than $1.0$ pixels. If less then $20$ reference points were kept, the pair was not used in the latter evaluation. 

We chose Graph-Cut RANSAC~\cite{barath2017graph} as a robust estimator since it can be considered as state-of-the-art and its source code is publicly available\footnote{\url{https://github.com/danini/graph-cut-ransac}}. In brief, it is a locally optimized RANSAC using graph-cut to achieve efficiency and global optimality w.r.t.\ the current so-far-the-best model. 

Validating the estimated fundamental matrices, we used the reference point sets. The geometric error was computed as the mean symmetric epipolar distance as follows: 
\begin{equation}
	\frac{1}{2} \sum_{(\Point_1, \Point_2) \in \mathcal{P}_{\text{R}}} \frac{\Fund \Point_1}{\sqrt{(\Fund \Point_1)^2_1 + (\Fund \Point_1)^2_2}} + \frac{\Point_2^\trans \Fund}{\sqrt{(\Point_2^\trans \Fund)^2_1 + (\Point_2^\trans \Fund)^2_2}},
	\label{eq:geometric_error}
\end{equation}
where $\mathcal{P}_{\text{R}}$ is the set of reference correspondences. 

The competitor methods, i.e.\ the minimal solvers combined with GC-RANSAC, were the normalized eight- and seven-point algorithms\footnote{OpenCV is used for the eight- and seven-point algorithms.}. In the least-squares model re-fitting step of GC-RANSAC, the normalized eight-point method was applied using the current inlier set.

Blocks (a--f) of Table~\ref{tab:Strecha_comparison} reports the mean result of $100$ runs on each pair from the {\fontfamily{cmtt}\selectfont Strecha} dataset. The first column is the name of the sequence, the second one is the number of the image pairs -- the ones having more than $20$ reference points. The next two blocks, each consisting of three columns, show the results of the methods if the confidence of GC-RANSAC is set to $99\%$ (1st block) and for a strict 30 FPS time limit (interrupted after $1 / 30$ secs; 2nd block). The reported properties are the geometric error of the estimated fundamental matrices (Eq.~\ref{eq:geometric_error}) w.r.t.\ the reference point sets, and the number of the samples drawn by GC-RANSAC. It can be seen that using the proposed method leads to \textit{more accurate model estimates using less samples} than the competitor algorithms. However, this test is slightly unfair since {\fontfamily{cmtt}\selectfont Strecha} dataset consists of images of buildings with large planar facades. Thus finding three co-planar points is not a challenging task.  

Blocks (g--i) show the mean results on {\fontfamily{cmtt}\selectfont AdelaideRMF}, {\fontfamily{cmtt}\selectfont Kusvod2} and {\fontfamily{cmtt}\selectfont Multi-H} datasets (1st column) if the confidence is set to $99\%$ (4th -- 6th cols) and for a strict $1 / 30$ seconds time limit (7th -- 9th cols). It can be seen that for both cases, the proposed method achieved the lowest mean errors in all but one test cases.   

\begin{table*}
\center
\caption{ Fundamental matrix estimation using GC-RANSAC~\cite{barath2017graph} with minimal methods ($2$nd row) applied to the sequences of the {\fontfamily{cmtt}\selectfont Strecha} dataset. The $1$st column shows the sequences: (a) Fountain-P11, (b) {\fontfamily{cmtt}\selectfont Entry-p10}, (c) {\fontfamily{cmtt}\selectfont Castle-p19}, (d) {\fontfamily{cmtt}\selectfont Castle-p30}, (e) Herzjesus-p8, and (f) {\fontfamily{cmtt}\selectfont Herzjesus-p25}, (g) {\fontfamily{cmtt}\selectfont Kusvod2}, (h) {\fontfamily{cmtt}\selectfont AdelaideRMF}, and (i) {\fontfamily{cmtt}\selectfont Multi-H}. The number of the image pairs and the tested properties are reported in the $2$nd and $3$rd columns. The next three report the results at $99\%$ confidence. For the remaining columns, there was a time limit set to $30$ FPS, i.e.\ the run is interrupted after $1 / 30$ secs. Values are the means of $100$ runs. The mean geometric error (in pixels) of the results w.r.t.\ the manually annotated inliers are written in each $1$st row; the required number of samples are reported in every $2$th row. The error is the symmetric epipolar distance. }
  	\begin{tabular}{| l | l | r || r | r | r || r | r | r |  }
    \hline
 	 	 \multicolumn{3}{|c||}{} & \multicolumn{3}{c||}{Confidence 99\%} & \multicolumn{3}{c|}{30 FPS} \\ 
    \hline 
 	 	 \multicolumn{3}{|r||}{Minimal methods $\to$} & \multicolumn{1}{c|}{\textbf{5}} & \multicolumn{1}{c|}{7} & \multicolumn{1}{c||}{8} & \multicolumn{1}{c|}{\textbf{5}} & \multicolumn{1}{c|}{7} & \multicolumn{1}{c|}{8} \\ 
    \hline 
 	 	\multirow{2}{*}{\rot{(a)}} & \multirow{2}{*}{\rot{53}} & Avg Err (px) & \textbf{3.06} & 4.34 & 16.21 & \textbf{4.31} & 7.29 & 17.15 \\
 	 	 & & Samples & \textbf{3\;692} & 5\;084 & 5\;471 & 42 & \textbf{38} & 59 \\
    \hline
 	 	\multirow{2}{*}{\rot{(b)}} & \multirow{2}{*}{\rot{45}} & Avg Err (px) & \textbf{1.42} & 1.63 & 3.10 & \textbf{2.33} & 3.93 & 8.95 \\
 	 	 & & Samples & \textbf{4\;953} & 6\;621 & 7\;045 & 40 & \textbf{36} & 57 \\
    \hline
 	 	\multirow{2}{*}{\rot{(c)}} & \multirow{2}{*}{\rot{81}} & Avg Err (px) & \textbf{6.71} & 9.52 & 20.54 & \textbf{6.80} & 10.75 & 23.92 \\
 	 	 & & Samples & \textbf{6\;450} & 7\;394 & 7\;586 & 30 & \textbf{29} & 33 \\
    \hline
 	 	\multirow{2}{*}{\rot{(d)}} & \multirow{2}{*}{\rot{196}} & Avg Err (px) & \textbf{5.40} & 8.71 & 20.51 & \textbf{6.78} & 8.82 & 19.01 \\
 	 	 & & Samples & \textbf{6\;720} & 7\;780 & 8\;094 & 49 & \textbf{42} & 82 \\
    \hline
 	 	\multirow{2}{*}{\rot{(e)}} & \multirow{2}{*}{\rot{26}} & Avg Err (px) & \textbf{2.86} & 6.08 & 19.85 & 7.36 & \textbf{6.54} & 19.38 \\
 	 	 & & Samples & \textbf{5\;432} & 6\;545 & 7\;088 & 45 & \textbf{40} & 74 \\
    \hline
 	 	\multirow{2}{*}{\rot{(f)}} & \multirow{2}{*}{\rot{114}} & Avg Err (px) & \textbf{4.84} & 9.14 & 16.21 & \textbf{7.69} & 10.06 & 27.83 \\
 	 	 & & Samples & \textbf{5\;881} & 7\;100 & 7\;434 & 58 & \textbf{47} & 103 \\
    \hline
 	 	\multirow{2}{*}{\rot{(g)}} & \multirow{2}{*}{\rot{18}} & Avg Err (px) & 0.63 & \textbf{0.52} & 0.53 & 0.70 & \textbf{0.56} & 0.59 \\
 	 	 & & Samples & \textbf{523} & 1\;178 & 1\;656 & \textbf{153} & 232 & 413 \\
    \hline
 	 	\multirow{2}{*}{\rot{(h)}} & \multirow{2}{*}{\rot{24}} & Avg Err (px) & \textbf{6.11} & 6.93 & 9.08 & \textbf{7.44} & 7.55 & 10.94  \\
 	 	 & & Samples & \textbf{1\;353} & 2\;273 & 2\;859 & \textbf{100} & 182 & 285 \\
    \hline
 	 	\multirow{2}{*}{\rot{(i)}} & \multirow{2}{*}{\rot{4}} & Avg Err (px) & \textbf{0.34} & 0.37 & 0.38 & \textbf{0.79} & 0.97 & 5.46 \\
 	 	 & & Samples & \textbf{1\;985} & 3\;299 & 4\;991 & 42 & \textbf{33} & 68 \\
    \hline
    \hline
 	 	\multirow{2}{*}{\rot{(all)}} & \multirow{2}{*}{\rot{561}} & Avg Err (px) & \textbf{3.47} & 7.41 & 16.53 & \textbf{4.90} & 8.33 & 19.51 \\
 	 	 & & Samples & \textbf{5\;560} & 6\;276 & 7\;055 & \textbf{52} & \textbf{52} & 93 \\
    \hline
\end{tabular}
\label{tab:Strecha_comparison}
\end{table*}

Fig.~\ref{fig:increasing_baseline} shows the error (in pixels) and the sample number plotted as the function of the baseline (in degrees). The results are the mean of $100$ runs on each image pair, \#515 on total, of the {\fontfamily{cmtt}\selectfont Strecha} dataset. Since the cameras in the sequences move around a building with approx. $180^{\circ}$, the baseline is indicated by the current angle.

Fig.~\ref{fig:real_images} shows example image pairs from each dataset with the epipolar lines of $50$ random inliers and five correspondences used as a minimal sample in the proposed method (red dots). It can be seen, that the results seem good: the epipolar lines go through the same pixels in the first (left) and second (right) images. Pairs (a) and (b) show an interesting effect: there are no entirely co-planar three points. 
Nevertheless, the initially estimated fundamental matrix was precise enough to be accurately refined by the local optimization step of GC-RANSAC. 

\subsection{Application: Rigid Motion Segmentation}

In this section, we show an possible application where estimating a fundamental matrix using fewer points than the state-of-the-art is beneficial. 

Multiple rigid motions in two views can be interpreted as a set of fundamental matrices. Typically, they are estimated by applying a multi-model fitting algorithm like PEARL~\cite{isack2012energy}. State-of-the-art fitting algorithms generate a set of initial fundamental matrices using a RANSAC-like sampling combined with a minimal method. Then an optimization is applied assigning the points to motion clusters and selecting the motions best interpreting the scene. 

\begin{figure}[htbp]
	\centering
    \begin{subfigure}{0.99\columnwidth}
		\includegraphics[width = 0.49\columnwidth]{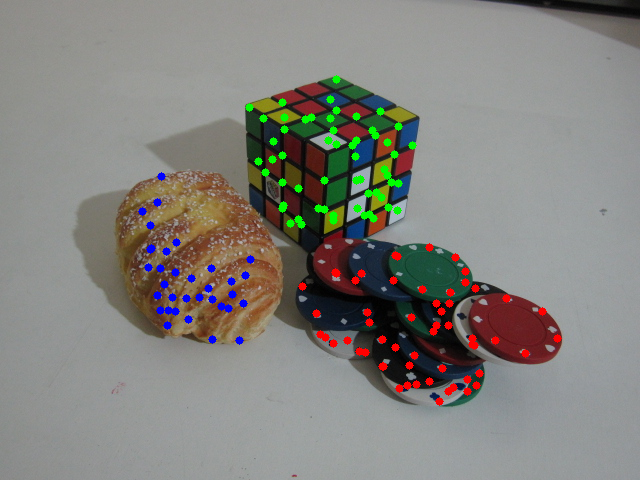}
		\includegraphics[width = 0.49\columnwidth]{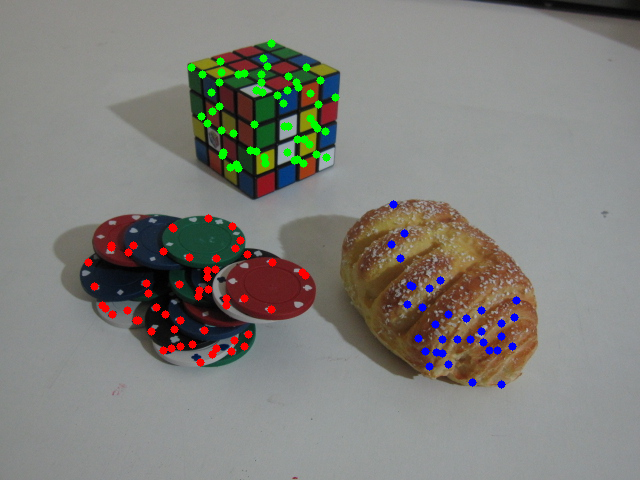}
		\caption{ breadcubechips }
    \end{subfigure}
    \begin{subfigure}{0.99\columnwidth}
		\includegraphics[width = 0.49\columnwidth]{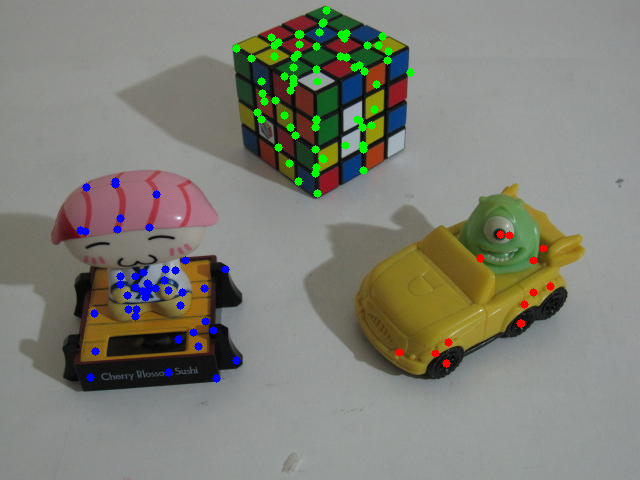}
		\includegraphics[width = 0.49\columnwidth]{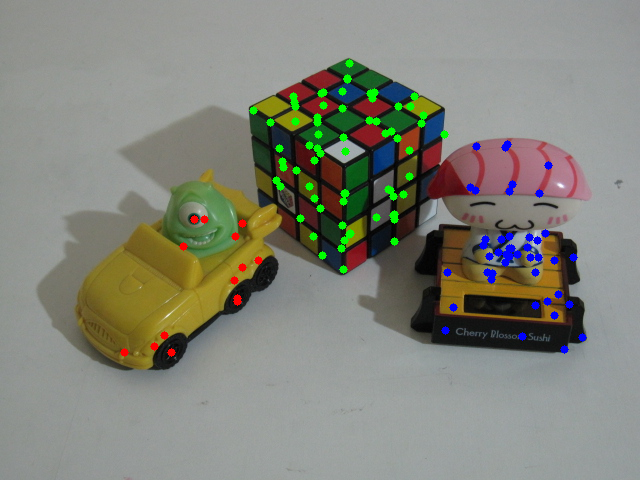}
		\caption{ toycubecar }
    \end{subfigure}
	\caption{ Example results of PEARL~\cite{isack2012energy} combined with the proposed algorithm applied to the {\fontfamily{cmtt}\selectfont {\fontfamily{cmtt}\selectfont AdelaideRMF}} motion dataset. Colors denote motions, black dots are outliers. }
	\label{fig:real_motion_images}
\end{figure}

The methods were evaluated on the {\fontfamily{cmtt}\selectfont AdelaideRMF} motion dataset (see Fig.~\ref{fig:real_motion_images} for examples) consisting of $18$ image pairs and the ground truth -- correspondences assigned to their motion clusters or outlier class.
Table~\ref{tab:motion_comparison} reports the result of PEARL combined with minimal methods (rows). The error is the misclassification error
\begin{equation*}
	\text{ME} = \frac{\# \text{Misclassified Points}}{\# \text{Points}},
\end{equation*}
which is the ratio of the points not assigned to the desired motion cluster. PEARL used the same initial model number for all methods, i.e.\ twice the input point number. The inlier-outlier threshold was tuned for each problem and each method separately. 
It can be seen that by using the five-point algorithm, the \textit{obtained clusterings are the most accurate}. 

\begin{table}
\center
\caption{ Two-view multi-motion clustering by PEARL~\cite{isack2012energy} combined with minimal methods (5th -- 7th cols). The number of points (P), motions (M) and the outlier percentage (O). The reported misclassification errors (in percentage) are the ratio of the points assigned to not the desired motion cluster. Test pairs from the {\fontfamily{cmtt}\selectfont {\fontfamily{cmtt}\selectfont AdelaideRMF}} motion dataset: (1) biscuit, (2) biscuitbookbox, (3) boardgame, (4) book, (5) breadcartoychips, (6) breadcube, (7) breadcubechips, (8) breadtoy, (9) breadtoycar, (10) carchipscube, (11) cube, (12) cubebreadtoychips, (13) cubechips, (14) cubetoy, (15) dinobooks, (16) game, (17) gamebiscuit, (18) toycubecar.  }
	\resizebox{.99\columnwidth}{!}{\begin{tabular}{| r | c c c | c c c | }
        \hline
    		& P & M & O & \textbf{5-point} & 7-point & 8-point \\
        \hline
            (1) & 330 & 1 & 55.8 & \phantom{x}1.4 & \phantom{x}2.7 & \phantom{x}\textbf{0.0} \\ 
            (2) & 259 & 3 & 37.5 & \phantom{x}\textbf{2.2} & \phantom{x}3.0 & \phantom{x}\textbf{2.2} \\ 
            (3) & 279 & 3 & 40.5 & \phantom{x}9.0 & \phantom{x}8.6 & \phantom{x}\textbf{7.5} \\ 
            (4) & 187 & 1 & 43.9 & \phantom{x}\textbf{2.1} & \phantom{x}\textbf{2.1} & \phantom{x}\textbf{2.1} \\ 
            (5) & 237 & 4 & 34.6 & \phantom{x}6.1 & \phantom{x}\textbf{5.1} & \phantom{x}7.0 \\ 
            (6) & 242 & 2 & 31.8 & \phantom{x}\textbf{1.5} & \phantom{x}\textbf{1.5} & \phantom{x}2.5 \\ 
            (7) & 230 & 3 & 35.2 & \phantom{x}\textbf{0.9} & \phantom{x}1.9 & \phantom{x}2.0 \\ 
            (8) & 288 & 2 & 36.8 & \phantom{x}1.4 & \phantom{x}\textbf{1.3} & \phantom{x}1.8 \\ 
            (9) & 166 & 3 & 33.7 & \phantom{x}7.9 & \phantom{x}\textbf{6.5} & \phantom{x}6.6 \\ 
            (10) & 165 & 3 & 36.4 & \phantom{x}3.8 & \phantom{x}3.8 & \phantom{x}\textbf{3.1} \\ 
            (11) & 302 & 1 & 67.9 & \textbf{10.4} & 12.4 & 12.4 \\ 
            (12) & 327 & 4 & 26.9 & \phantom{x}\textbf{3.2} & \phantom{x}3.8 & \phantom{x}4.1 \\ 
            (13) & 284 & 2 & 50.3 & \phantom{x}\textbf{1.6} & \phantom{x}3.8 & \phantom{x}4.9 \\ 
            (14) & 249 & 2 & 39.8 & \phantom{x}1.0 & \phantom{x}\textbf{0.5} & \phantom{x}1.6 \\ 
            (15) & 360 & 3 & 43.1 & 15.1 & 16.3 & \textbf{10.5} \\ 
            (16) & 233 & 1 & 73.0 & \phantom{x}\textbf{5.5} & \phantom{x}6.1 & \phantom{x}\textbf{5.5} \\ 
            (17) & 328 & 2 & 50.9 & \phantom{x}\textbf{1.6} & \phantom{x}2.4 & \phantom{x}2.2 \\ 
            (18) & 200 & 3 & 36.0 & \phantom{x}5.4 & \phantom{x}7.1 & \phantom{x}\textbf{4.3} \\ 
        \hline
            Avg & & & & \phantom{x}\textbf{4.5} & \phantom{x}4.9 & \phantom{x}\textbf{4.5} \\ 
            Med & & & & \phantom{x}\textbf{2.7} & \phantom{x}3.8 & \phantom{x}3.6 \\ 
        \hline
\end{tabular}}
\label{tab:motion_comparison}
\end{table}
 
\subsection{Processing Time}

The proposed method consists of three main steps: (i) the null-space computation of a matrix of size $6 \times 9$, then the homography parameters are calculated in closed form. (ii) Using the estimated $\Hom$ and two additional correspondences, a coefficient matrix of size $7 \times 9$ is built and its null-space is computed. (iii) Finally, the roots of a cubic polynomial are estimated. The average processing time of $100$ runs of our C++ implementation using OpenCV was $0.16$ milliseconds. 

Combining RANSAC-like \textit{hypothesize-and-verify} robust estimators with the proposed method is beneficial since their processing time highly depends on the size of the minimal sample required for the estimation. Table~\ref{tab:ransac_iterations} shows the theoretically needed iteration number of RANSAC combined with minimal methods (columns) on different outlier levels (rows). The confidence value was set to $95\%$. It can be seen that using the proposed 5-point algorithm leads to significant improvement in the processing time. 

\begin{table}[h!]
  \centering
  \caption{Required theoretical iteration number of RANSAC~\cite{fischler1981random} combined with minimal methods (columns) with confidence set to $95\%$ on different outlier levels (rows).}
  \begin{tabular}{ | c | r r r | }
  \hline
      & \multicolumn{3}{c|}{Confidence $95\%$}\\ 
  \hline
    Outl. & \textbf{5}{\footnotesize\phantom{xx}} & 7{\footnotesize\phantom{xx}} & 8{\footnotesize\phantom{xx}}  \\ 
  \hline
    50\% & $\mathbf{94}${\footnotesize\phantom{xx}} & $382${\footnotesize\phantom{xx}} & $765${\footnotesize\phantom{xx}} \\ 
    80\% & $\mathbf{\sim10^4}${\footnotesize\phantom{x}} & $\sim10^5${\footnotesize\phantom{x}} & $\sim10^6${\footnotesize\phantom{x}} \\
    95\% & $\mathbf{\sim10^7}${\footnotesize\phantom{x}} & $\sim10^9${\footnotesize\phantom{x}} & $\sim10^{10}$ \\
    99\% & $\mathbf{\sim10^{10}}$ & $\sim10^{14}$ & $\sim10^{16}$\\
  \hline
  \end{tabular}
  \label{tab:ransac_iterations}
\end{table}

\section{Conclusion}

In this paper, we proposed a method for estimating the fundamental matrix between two non-calibrated cameras from five correspondences of rotation invariant features. Three of the points have to be co-planar and two of them be in general position. The solver, combined with Graph-Cut RANSAC, was superior to the seven- and eight-point algorithms both in terms of accuracy and needed sample number on the evaluated $561$ publicly available real image pairs.  
It is demonstrated that the co-planarity of three points is not a too restrictive constraint in real world (e.g.\ in urban environment) and can be weakened by state-of-the-art robust estimators. 
Moreover, we showed that the method makes multi-motion fitting more accurate than using the eight- or seven-point algorithms. 
%

\appendix 

\section{Calculation of the Homography Parameters}
\label{app:homography_parameters}

In this section, we show how parameters $\beta$ and $\gamma$ in Eqs.~\ref{eq:null_space} are calculated. Replacing each $h_j$ with $\beta b_j + \gamma c_j + d_j$ ($j \in [1,9]$) in the $1$st and $3$rd equations of Eqs.~\ref{eq:hom_aff_sys} leads to the following system:
\begin{eqnarray*}
          (\beta b_1 + \gamma c_1 + d_1) - u_2 (\beta b_7 + \gamma c_7 + d_7) - \\
          u_1 c_{\alpha} s_{u} (\beta b_7 + \gamma c_7 + d_7) - \\
          v_1 c_{\alpha} s_{u} (\beta b_8 + \gamma c_8 + d_8) - c_{\alpha} s_{u} = 0, \\
		(\beta b_4 + \gamma c_4 + d_4) - v_2 (\beta b_7 + \gamma c_7 + d_7) - \\
		u_1 s_{\alpha} s_{u} (\beta b_7 + \gamma c_7 + d_7) - \\
		v_1 s_{\alpha} s_{u} (\beta b_8 + \gamma c_8 + d_8) - s_{\alpha} s_{u} = 0. 
\end{eqnarray*}
After expanding and rearranging the expressions, the first equation becomes
\begin{eqnarray*}
        (b_1 - u_2 b_7) \beta + (c_1 - u_2 c_7) \gamma - 
        (u_1 c_{\alpha} b_7 + v_1 c_{\alpha} b_8) s_{u} \beta - \\ 
        (u_1 c_{\alpha} d_7 + v_1 c_{\alpha} d_8 + c_{\alpha}) s_{u} + 
        (u_1 c_{\alpha} c_7 + v_1 c_{\alpha} c_8) s_{u} \gamma - \\
        d_1 - u_2 d_7  = 0,
\end{eqnarray*}
and the second one is as follows: 
\begin{eqnarray*}
		(b_4 - v_2 b_7) \beta + (c_4 - v_2 c_7) \gamma - (u_1 s_{\alpha} b_7 + v_1 s_{\alpha} b_8) s_{u} \beta - \\
		(u_1 s_{\alpha} d_7 + v_1 s_{\alpha} d_8 + s_{\alpha}) s_{u} - (u_1 s_{\alpha} c_7 + v_1 s_{\alpha} c_8) s_{u} \gamma - \\
		d_4 - v_2 d_7 = 0. 
\end{eqnarray*}
The monomials of this polynomial system are $[\beta \quad \gamma \quad s_u \quad s_{u} \beta \quad s_u \gamma ]^\trans$. 

Having two rotations $\alpha_1$ and $\alpha_2$ doubles the equations and introduces another unknown (each correspondence has different $s_u$). Thus the monomials of the polynomial equation system to which the two rotations lead are $[\beta \quad \gamma \quad s_{u,1} \quad s_{u,1} \beta \quad s_{u,1} \gamma \quad s_{u,2} \quad s_{u,2} \beta \quad s_{u,2} \gamma]^\trans$, where $s_{u,i}$ is the scale along axis $u$ of the $i$th correspondence ($i \in \{1,2\}$). Since four equations are given for four unknowns and there is no higher-order term, the system can straightforwardly be rearranged and solved. The formulas for $\beta$ and $\gamma$ are as follows:
\begin{equation*}
  \begin{array}{ll}
      \beta = & (-c_{\alpha_2} c_1 d_7 v_{2,2} s_{\alpha_1} + c_{\alpha_2} c_4 d_7 u_{2,1} s_{\alpha_1} + c_{\alpha_2} c_7 d_1 v_{2,2} s_{\alpha_1} \\
       & - c_{\alpha_2} c_7 d_4 u_{2,1} s_{\alpha_1} - c_{\alpha_2} c_{\alpha_1} c_4 d_7 v_{2,1} + c_{\alpha_2} c_{\alpha_1} c_4 d_7 v_{2,2} \\
       & + c_{\alpha_2} c_{\alpha_1} c_7 d_4 v_{2,1} - c_{\alpha_2} c_{\alpha_1} c_7 d_4 v_{2,2} - c_1 d_7 u_{2,1} s_{\alpha_2} s_{\alpha_1} \\
       & + c_1 d_7 u_{2,2} s_{\alpha_2} s_{\alpha_1} + c_7 d_1 u_{2,1} s_{\alpha_2} s_{\alpha_1} - c_7 d_1 u_{2,2} s_{\alpha_2} s_{\alpha_1} \\
       & + c_{\alpha_1} c_1 d_7 v_{2,1} s_{\alpha_2} - c_{\alpha_1} c_4 d_7 u_{2,2} s_{\alpha_2} - c_{\alpha_1} c_7 d_1 v_{2,1} s_{\alpha_2} \\
       & + c_{\alpha_1} c_7 d_4 u_{2,2} s_{\alpha_2} + c_{\alpha_2} c_1 d_4 s_{\alpha_1} - c_{\alpha_2} c_4 d_1 s_{\alpha_1} \\
       & - c_{\alpha_1} c_1 d_4 s_{\alpha_2} + c_{\alpha_1} c_4 d_1 s_{\alpha_2}) /  \\
       & (c_{\alpha_2} b_1 c_7 v_{2,2} s_{\alpha_1} + c_{\alpha_2} b_4 c_7 u_{2,1} s_{\alpha_1} + c_{\alpha_2} b_7 c_1 v_{2,2} s_{\alpha_1} \\
       & - c_{\alpha_2} b_7 c_4 u_{2,1} s_{\alpha_1} - c_{\alpha_2} c_{\alpha_1} b_4 c_7 v_{2,1} + c_{\alpha_2} c_{\alpha_1} b_4 c_7 v_{2,2} \\
       & + c_{\alpha_2} c_{\alpha_1} b_7 c_4 v_{2,1} - c_{\alpha_2} c_{\alpha_1} b_7 c_4 v_{2,2} - b_1 c_7 u_{2,1} s_{\alpha_1} s_{\alpha_2} \\
       & + b_1 c_7 u_{2,2} s_{\alpha_1} s_{\alpha_2} + b_7 c_1 u_{2,1} s_{\alpha_1} s_{\alpha_2} - b_7 c_1 u_{2,2} s_{\alpha_1} s_{\alpha_2} \\
       & + c_{\alpha_1} b_1 c_7 v_{2,1} s_{\alpha_2} - c_{\alpha_1} b_4 c_7 u_{2,2} s_{\alpha_2} - c_{\alpha_1} b_7 c_1 v_{2,1} s_{\alpha_2} \\
       & + c_{\alpha_1} b_7 c_4 u_{2,2} s_{\alpha_2} + c_{\alpha_2} b_1 c_4 s_{\alpha_1} - c_{\alpha_2} b_4 c_1 s_{\alpha_1} \\
       & - c_{\alpha_1} b_1 c_4 s_{\alpha_2} + c_{\alpha_1} b_4 c_1 s_{\alpha_2}), 
  \end{array}
\end{equation*}
\begin{equation*}
  \begin{array}{ll}
      \gamma = & - (-c_{\alpha_2} b_1 d_7 v_{2,2} s_{\alpha_1} + c_{\alpha_2} b_4 d_7 u_{2,1} s_{\alpha_1} + c_{\alpha_2} b_7 d_1 v_{2,2} s_{\alpha_1} \\
         & - c_{\alpha_2} b_7 d_4 u_{2,1} s_{\alpha_1} - c_{\alpha_2} c_{\alpha_1} b_4 d_7 v_{2,1} + c_{\alpha_2} c_{\alpha_1} b_4 d_7 v_{2,2} \\
         & + c_{\alpha_2} c_{\alpha_1} b_7 d_4 v_{2,1} - c_{\alpha_2} c_{\alpha_1} b_7 d_4 v_{2,2} - b_1 d_7 u_{2,1} s_{\alpha_1} s_{\alpha_2} \\
         & + b_1 d_7 u_{2,2} s_{\alpha_1} s_{\alpha_2} + b_7 d_1 u_{2,1} s_{\alpha_1} s_{\alpha_2} - b_7 d_1 u_{2,2} s_{\alpha_1} s_{\alpha_2} \\
         & + c_{\alpha_1} b_1 d_7 v_{2,1} s_{\alpha_2} - c_{\alpha_1} b_4 d_7 u_{2,2} s_{\alpha_2} - c_{\alpha_1} b_7 d_1 v_{2,1} s_{\alpha_2} \\
         & + c_{\alpha_1} b_7 d_4 u_{2,2} s_{\alpha_2} + c_{\alpha_2} b_1 d_4 s_{\alpha_1} - c_{\alpha_2} b_4 d_1 s_{\alpha_1} \\
         & - c_{\alpha_1} b_1 d_4 s_{\alpha_2} + c_{\alpha_1} b_4 d_1 s_{\alpha_2}) / \\
         & (- c_{\alpha_2} b_1 c_7 v_{2,2} s_{\alpha_1} + c_{\alpha_2} b_4 c_7 u_{2,1} s_{\alpha_1} + c_{\alpha_2} b_7 c_1 v_{2,2} s_{\alpha_1} \\
         & - c_{\alpha_2} b_7 c_4 u_{2,1} s_{\alpha_1} - c_{\alpha_2} c_{\alpha_1} b_4 c_7 v_{2,1} + c_{\alpha_2} c_{\alpha_1} b_4 c_7 v_{2,2} \\
         & + c_{\alpha_2} c_{\alpha_1} b_7 c_4 v_{2,1} - c_{\alpha_2} c_{\alpha_1} b_7 c_4 v_{2,2} - b_1 c_7 u_{2,1} s_{\alpha_1} s_{\alpha_2} \\
         & + b_1 c_7 u_{2,2} s_{\alpha_1} s_{\alpha_2} + b_7 c_1 u_{2,1} s_{\alpha_1} s_{\alpha_2} - b_7 c_1 u_{2,2} s_{\alpha_1} s_{\alpha_2} \\
         & + c_{\alpha_1} b_1 c_7 v_{2,1} s_{\alpha_2} - c_{\alpha_1} b_4 c_7 u_{2,2} s_{\alpha_2} - c_{\alpha_1} b_7 c_1 v_{2,1} s_{\alpha_2} \\
         & + c_{\alpha_1} b_7 c_4 u_{2,2} s_{\alpha_2} + c_{\alpha_2} b_1 c_4 s_{\alpha_1} - c_{\alpha_2} b_4 c_1 s_{\alpha_1} \\
         & - c_{\alpha_1} b_1 c_4 s_{\alpha_2} + c_{\alpha_1} b_4 c_1 s_{\alpha_2}).
  \end{array}
\end{equation*}

{\small
\bibliographystyle{ieee}
\bibliography{egbib}
}

\end{document}